\documentclass[journal]{IEEEtran}

\usepackage[utf8]{inputenc}
\usepackage{amsmath,amssymb}
\usepackage{graphicx}
\usepackage{siunitx}
\usepackage{url}
\usepackage{cite}
\usepackage{booktabs}
\usepackage{multirow}
\usepackage[table,xcdraw]{xcolor}
\usepackage{hyperref}

\usepackage{tikz}
\usepackage{textcomp}
\usepackage[doipre={DOI:~}]{uri}
\usepackage{lipsum}

\usepackage[absolute,showboxes]{textpos}

\TPMargin{5pt}

\newcommand{\copyrightstatement}{
    \begin{textblock}{0.92}(0.04,0.96)
         \noindent
         \scriptsize \textcopyright 2026 IEEE. Personal use of this material is permitted.  Permission from IEEE must be obtained for all other uses, in any current or future media, including reprinting/republishing this material for advertising or promotional purposes, creating new collective works, for resale or redistribution to servers or lists, or reuse of any copyrighted component of this work in other works.
    \end{textblock}
}

\begin{document}
\bstctlcite{IEEEexample:BSTcontrol}

\title{MAUPITI: On-Device Prototype-Based Learning on a Smart Infrared Sensor}

\author{Beatrice Alessandra Motetti, Tanguy Dugas du Villard, Matteo Risso, Alessio Burrello, Francesco Daghero, Enrico Macii, Massimo Poncino, Marco Castellano, Alfio Basile, Daniele Jahier Pagliari%

\thanks{B.A. Motetti, T. Dugas du Villard, M. Risso, A. Burrello, F. Daghero,
E. Macii, M. Poncino, and D. Jahier Pagliari are with Politecnico di Torino, Turin, Italy. Corresponding author email: beatrice.motetti@polito.it.

M. Castellano, and A. Basile are with STMicroelectronics, Cornaredo, Italy.

This publication is part of the project PNRR-NGEU which has received funding from the MUR – DM 118/2023.}}

\maketitle
\copyrightstatement

\begin{abstract}
Low-resolution infrared (IR) array sensors represent an interesting solution for privacy-preserving human sensing in embedded systems. In this letter, we describe a smart multi-pixel IR sensor integrating a 16$\times$16 thermal MOSFET (TMOS) array and a RISC-V microcontroller extended with low-precision SIMD instructions, capable of \emph{on-device learning} and \emph{continual adaptation} for pose and gesture recognition tasks under tight memory and power constraints ($<$32\,kB on-chip memory, $\approx$1.5\,mW). To avoid the memory overheads of backpropagation and replay buffers, we adopt a prototype-based Nearest Class Mean (NCM) classifier in which a simple Convolutional Neural Network (CNN) encoder is trained and quantized offline, while class prototypes are stored and updated on the device in streaming mode. With experiments on two datasets, we show that this approach yields accuracy on par with a conventional classifier, with negligible latency overheads in both the classification and the prototype update ($<$0.29\% considering both phases), effectively enabling online adaptation of the perception framework. 
\end{abstract}


\section{Introduction and Related Works}
\label{sec:intro}

Low-resolution infrared (IR) arrays are increasingly adopted in embedded systems for occupancy monitoring, people counting, and human-computer interaction, as they combine low power consumption, low cost, and strong privacy guarantees by sensing only coarse thermal patterns instead of high-resolution RGB images \cite{XieIoTJ2023, Bouazizi2022}. 

Prior work has focused on exploring machine-learning-based algorithms for IR-based perception~\cite{diaz2024, yin2023har}. 
However, the focus of these early approaches was not on the direct deployment of the system on an embedded device, requiring expensive data transmissions to servers for inference~\cite{dlwithedge}. 
More recently, other works implemented IR perception algorithms able to run directly on embedded devices, such as a Raspberry PI~\cite{hybrid2021, kraft2021} or an STM32F microcontroller (MCU)~\cite{Metwaly2020,XieIoTJ2023}.
In our previous work \cite{RissoDATE2024}, we proposed a full-stack HW-SW optimization flow for tiny Convolutional Neural Networks (CNNs) on IR arrays, focusing on privacy-preserving people counting. By combining hardware-aware differentiable neural architecture search and mixed-precision quantization we produced models that could run on MAUPITI, a smart IR sensor integrating a 16$\times$16 TMOS array and a \emph{Ibex} RISC-V core.
By executing optimized CNNs directly on the sensor, we demonstrated significant energy benefits w.r.t. MCU-based deployments~\cite{RissoDATE2024}.
All these previous works, however, focused solely on \textit{local inference}, while training was kept offline on a more powerful machine.

In many smart-sensing deployments, fixed models are not sufficient. Sensor placement, background temperature, and patterns to recognize vary across locations and over time. Sending data to the cloud to re-train the model requires constant connectivity and may undermine energy efficiency. Instead, \emph{on-device learning} and \emph{continual adaptation} are desirable, but remain challenging under the tight memory and compute budgets of sensor-class embedded hardware.

\begin{figure}[t]
    \centering
    \includegraphics[trim={0cm 0cm 0cm 0.25cm},clip,width=\linewidth]{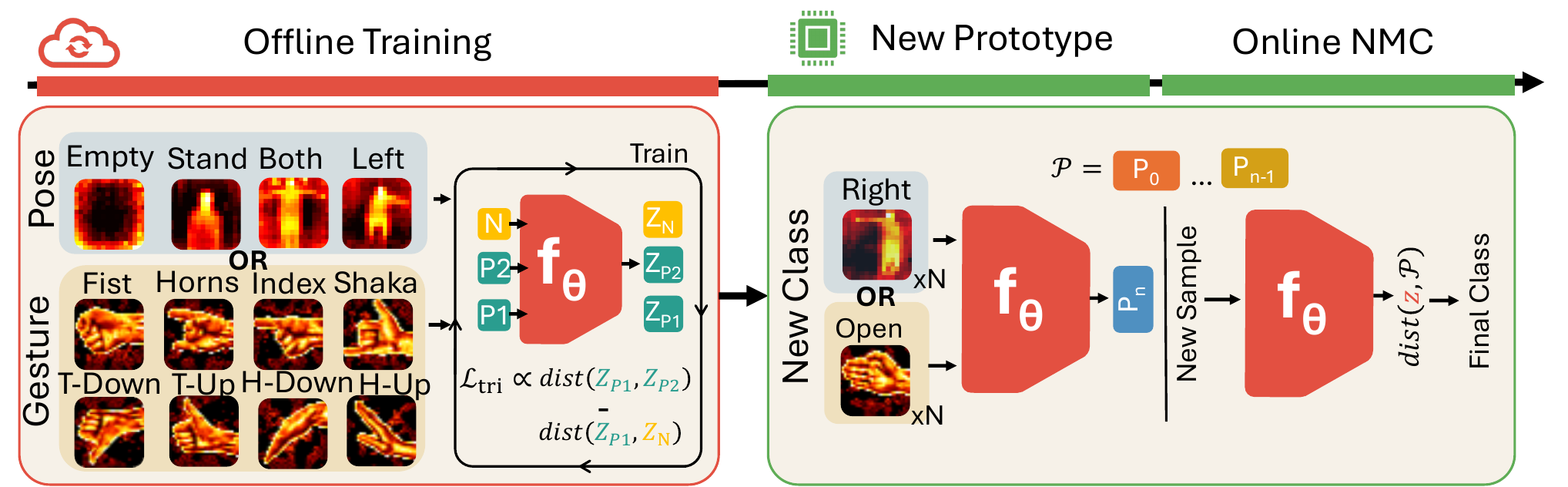}
    \caption{Overview of our proposed approach. An offline training is performed with triplet loss prior to deployment (left). New classes can easily be added online (right), with minimal overhead and without retraining the encoder.}
    \label{fig:method}
    \vspace{-.175cm}
\end{figure}

This letter extends MAUPITI~\cite{RissoDATE2024} from an inference-only smart sensor to a platform capable of \emph{on-device prototype-based learning}, as shown in Fig.~\ref{fig:method}. Namely:
\begin{itemize}
  \item We integrate a Nearest Class Mean (NCM) classifier\cite{SnellProtoNets, MensinkNCM} on top of a CNN encoder running on MAUPITI. The encoder is trained offline using metric learning and quantization-aware training, while prototypes can be updated on the device without backpropagation.
  \item We present an embedded implementation that fits within MAUPITI's 16\,kB instruction and 16\,kB data memories, enabling online prototype updates with negligible latency and memory overhead.
  \item We demonstrate the approach on two new applications for MAUPITI, i.e., pose and gesture recognition, achieving comparable accuracy with respect to a traditional CNN classifier trained with backpropagation, while enabling continual learning and few-shot adaptation on-board.
\end{itemize}

Our code and datasets are publicly available at  \href{https://github.com/eml-eda/maupiti-odl}{\texttt{https://github.com/eml-eda/maupiti-odl}}.


\section{Methodology}
\label{sec:methodology}

\subsection{Nearest Class Mean Classifiers}

Let $f_\theta(\cdot)$ denote the CNN encoder with parameters $\theta$, which we derive in this work by removing the classifier head from the CNN templates of~\cite{RissoDATE2024, XieIoTJ2023}. Given an input IR frame $\mathbf{x} \in \mathbb{R}^{16 \times 16}$, the encoder outputs an embedding
$
  \mathbf{z} = f_\theta(\mathbf{x}) \in \mathbb{R}^p.
$
\noindent where $p$ is the latent representation dimension (and prototype size).
In an NMC classifier~\cite{MensinkNCM}, for each class $c$, a prototype vector $\mathbf{m}_c \in \mathbb{R}^p$ is computed as the mean of the embeddings of all samples of that class ($\mathbf{x} \in \mathcal{D}_c$) encountered during training:
\begin{equation}
  \mathbf{m}_c = \frac{1}{|\mathcal{D}_c|} \sum_{i \in \mathcal{D}_c} \mathbf{z}_i
\end{equation}

At inference time, the distance between a new sample $\mathbf{x}$ and all prototypes is computed, and the class of the closest one in the embedding space corresponds to the NCM's prediction $\hat{c}$:
\begin{equation}
  \hat{c} = \arg\min_{c} d(\mathbf{z}, \mathbf{m}_c),
\end{equation}

\noindent where $d(\cdot,\cdot)$ is a distance metric (e.g., Euclidean or Manhattan). 
On-device training can be performed by incrementally updating prototypes. In particular, when a new labeled sample $(\mathbf{x}, c)$ is available, we derive $\mathbf{z} = f_\theta(\mathbf{x})$ and update the prototype of class $c$ accordingly:
\begin{equation}\label{eq:update}
  N_c \leftarrow N_c + 1,\ \ \ 
  \mathbf{m}_c \leftarrow \mathbf{m}_c + \frac{1}{N_c} \big( \mathbf{z} - \mathbf{m}_c \big).
\end{equation}
In a Continual Learning scenario, it is also straightforward to add a new class, as it is sufficient to allocate a new prototype slot and compute the mean latent representation over a set of new labeled samples, as shown in Fig.~\ref{fig:method}.

The prototype update has $O(p)$ time and memory complexity and does not require storing old samples in \textit{replay buffers}, nor gradients, making it suitable for sensors with a few tens of KiB of memory. Since no backpropagation is performed on-device, training latency is determined by embedding computation and prototype updates, with the latter adding only a few hundred operations on top of the forward pass of the CNN for realistic values of $p$. For more details on NCM models, we refer readers to~\cite{MensinkNCM}.

\subsection{Training Flow}
\label{sec:training_flow}
Our offline training (Fig.~\ref{fig:method}, left) comprises two stages:

\textbf{1) Metric learning} trains the encoder to generate an embedding space in which samples from the same class are clustered together, while samples of different classes are pushed away. We evaluated triplet loss with margins of 20, 50, and 100, as well as prototypical loss~\cite{SnellProtoNets}, using both Manhattan and Euclidean distances, and selected the Euclidean-distance triplet loss with a margin of 100 as it yielded the best results.

\textbf{2) Quantization-aware training (QAT)} to make the network compatible with the integer-only MAUPITI hardware.  We quantize using the PLiNIO library~\cite{PagliariPlinio}, employing INT8 affine quantization for weights and activations with PACT activation clipping~\cite{ChoiPACT}.

We also experimented with adding an initial pretraining stage in which the CNN encoder was trained in a supervised way, using a cross-entropy loss with a classification head that was later removed, but we experimentally found the performance of the NCM
classifier not to improve. Thus, the results in the paper are obtained by performing metric learning and QAT only.
After training, quantized models are compiled from PyTorch to C code using optimized kernel routines for MAUPITI, using a custom translator.

We explore the CNN encoder architecture through Bayesian optimization with Optuna~\cite{optuna_2019} varying the number of filters of the encoder's convolutional layers, along with the kernel size and the pooling size. The optimization is conducted with a dual objective, using Optuna's built-in multi-objective optimization capabilities: (i) maximizing the validation accuracy of the model on all classes on metric learning and (ii) minimizing the number of encoder parameters to reduce its memory occupation and fit MAUPITI's tight constraints.

\subsection{Embedded Deployment on MAUPITI}
MAUPITI is a smart infrared sensor SoC implemented in 130\,nm CMOS technology and clocked at 20\,MHz. It integrates two main blocks \cite{RissoDATE2024}:
\begin{itemize}
  \item A 16$\times$16 TMOS thermal array with 8 parallel analog front-ends, each reading one row, enabling acquisition of a full frame in two steps. The array draws approximately 0.62\,mW at 2.4\,V and reaches 10\,FPS.
  \item A digital processing block with a customized 32-bit RISC-V Ibex core, 16\,kB instruction memory, 16\,kB data memory, boot ROM, calibration registers, an instruction tracer, one-time programmable memory, and I2C/SPI3 communication interfaces. It also comprises an SDOTP unit that supports SIMD dot products between two 32-bit registers, interpreted as 4$\times$8-bit or 8$\times$4-bit vectors, plus a 32-bit accumulator. The digital block consumes about 0.9\,mW under nominal conditions.
\end{itemize}

In this work, we store prototypes on MAUPITI as INT32 vectors. For $C$ classes, prototype dimension $p$, the total prototype memory footprint is:
\begin{equation}
  M_{\text{proto}} = C \cdot p \cdot 4\;\text{bytes}
\end{equation}
Deployed models use custom hand-written kernels and a lightweight runtime. Convolutions and dense layers use INT8 inputs and weights. As our goal is to demonstrate the feasibility of NCM learning on a typical embedded sensor node, we do not use the SDOTP unit and instead perform scalar MAC operations with the main core, which are more representative of the typical hardware available on these devices.
Intermediate activations are stored in a pair of alternating buffers to minimize memory occupation while enabling concurrent inference and prototype updates. The NCM head reuses the same buffers to compute distances between the current embedding and prototypes. Moreover, we replace the division in Eq.~\ref{eq:update} with a shift by accumulating $\mathbf{z}$ values until $N_c$ is a power of two before subtracting and normalizing.

Online training is integrated into the same embedded C code that manages sensor acquisition and classification. Training is triggered by a control signal (e.g., a button press or a command from a host) specifying whether to create a new class or update an existing one.

\section{Experimental Evaluation}
\label{sec:experiments}

We perform most of our experiments on a simple \textbf{pose recognition} dataset, containing IR images of a human in front of the sensor in five different poses (\textit{empty scene}, \textit{standing}, \textit{both arms raised}, \textit{right arm raised}, and \textit{left arm raised}). To assess the generality of our approach, we additionally evaluate it on a more complex, 9-class \textbf{hand gesture recognition} dataset, including images of the following gestures, performed with the hand in front of the sensor, pointing sideways: \textit{fist}, \textit{horns}, \textit{index}, \textit{shaka}, \textit{thumb-down/up} \textit{wrist-down/up}, \textit{open hand}.
Both datasets are recorded in multiple indoor environments (offices, labs), with varied backgrounds and environmental temperatures. We split data session-wise as in~\cite{RissoDATE2024} to better match a real-world scenario in which the sensor is trained in a different environment w.r.t. the one in which it is deployed. 

Unless stated otherwise, we consider as encoder a tiny CNN that follows the blueprint architecture of~\cite{RissoDATE2024}, and includes three 2D-convolutional layers with 3x3 filters and 8 channels, followed by batch normalization and ReLU activations. Max pooling with 2x2 stride is inserted after the 2nd and 3rd convolution. Lastly, a single linear layer projects the flattened output of the last convolution to the prototype space of size $p$. 

In the following experiments we compare with three different baselines, considering a conventional CNN classifier with the same backbone, namely: i) full-network retraining, similarly to~\cite{castro2018finetuning}, ii) retraining only the last layer, as in previous works~\cite{pellegrini2020latentreplay, mei2024train_on_request}, and iii) retraining only the parameters associated to the added output neuron(s)~\cite{ren2021tinyol}. For each baseline we consider two variants, i.e., with and without replay.

\subsection{Pose recognition performance}

As a first experiment, we evaluate a NCM classifier in an offline training setting (all classes available in the initial training set), varying the prototype size $p$.
We obtain test accuracies in the range (91.72\%-92.76\%) for $p\!\in\!\{8, 16, 32, 64\}$, with a drop $\approx2-3\%$ relative to the 94.81\% obtained by the same CNN backbone attached to a softmax classifier head with 5 output neurons, thus confirming that replacing softmax with NCM does not significantly degrade accuracy for this task.

Next, we consider a continual learning setup, where either the last (1C) or the last two classes (2C) are not present during the initial training of the encoder and prototypes. These classes are introduced later on-device using a limited number of labeled samples (shots).
For this experiment we fix $p=64$, and we compare our approach with the baselines. When considering the variant with replay, we build the replay buffer by storing a random subset of samples from all other classes equal to the number of shots.

\begin{figure}[t]
    \centering
    \includegraphics[trim={0cm 0 0cm 0cm},clip,width=\linewidth]{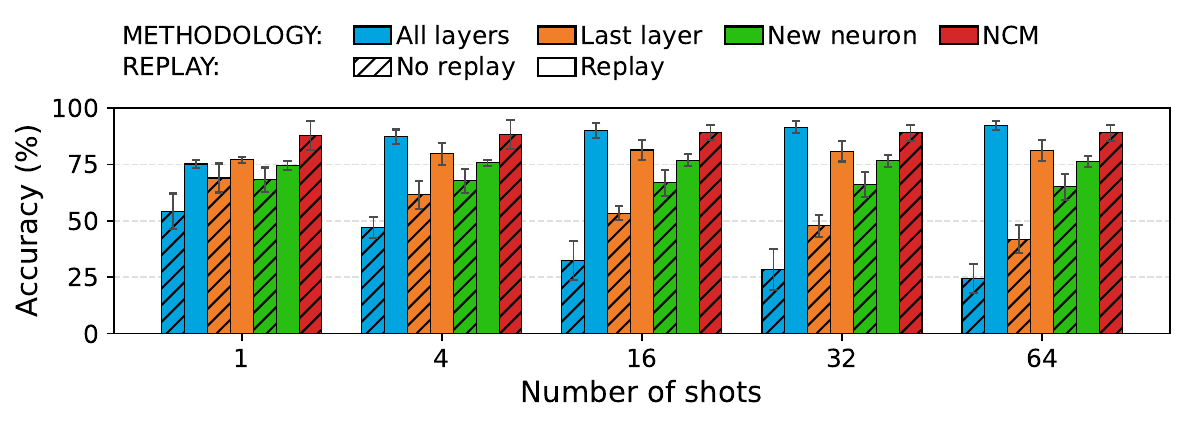}
    \includegraphics[trim={0cm 0 0cm 0cm},clip,width=\linewidth]{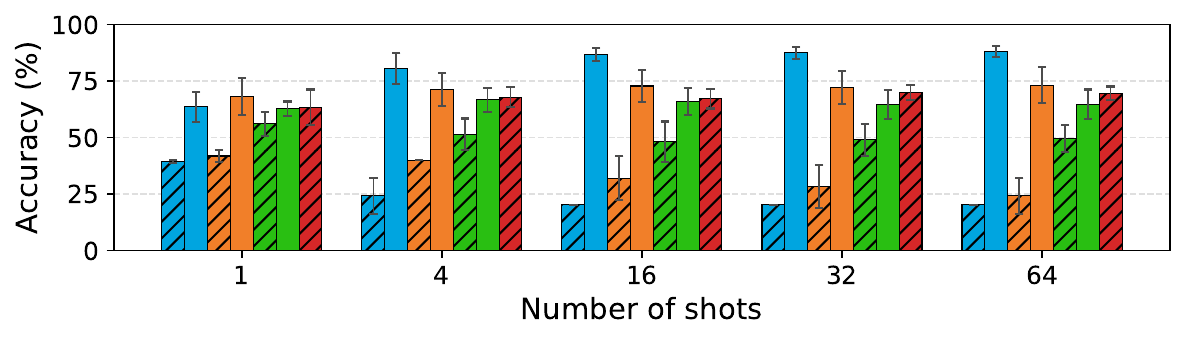}
    \vspace{-0.75cm}
    \caption{5-class accuracy of the NCM classifier and the considered baselines, trained online on one (top) or two (bottom) previously unseen classes. The error bars report the standard deviation over 5 seeds.}
    \label{fig:results_shots}
\end{figure}

Fig.~\ref{fig:results_shots} shows the accuracy results versus the number of shots for NCM and the considered baselines after the metric learning stage. We use power of two values for the number of shots because these allow replacing divisions with shifts as explained in Sec.~\ref{sec:methodology}. First, we note that NCM achieves very good accuracy even with a single sample to build the new class prototype, although the variability of results is reduced with 4+ shots. With 4 shots, the 1C average accuracy is 88\% (vs 94\% of offline training), which degrades to 68\% for 2C, as expected due to the increased difficulty of the online learning task without any encoder adaptation, as novel classes must be accommodated within a fixed latent space. In an additional scalability evaluation, NCM outperforms the best non-replay baseline by $\sim$4 percentage points in a 3C scenario.
Second, NCM significantly \textit{outperforms all other methods that do not use replay examples}, which suffer in particular in multiple-shots training due to catastrophic forgetting~\cite{shi2021}, achieving accuracies in the range (87.77\%, 89.04\%) for 1C and (63.40\%, 69.66\%) for 2C. This is despite NCM not incurring the large memory and latency/energy overheads of backpropagation, and being fully compatible with integer quantization, which is problematic for methods that require small weight updates. In the 1C case, NCM remains competitive even against replay methods, \textit{including full network fine-tuning}, albeit not requiring any extra input storage.

Next, we apply the Bayesian Optimization procedure described in Sec.~\ref{sec:training_flow} on the metric learning stage.
We obtain a Pareto front of CNNs, and report in Tab.~\ref{tab:results_bayopt} the 4-shots test accuracy after the QAT phase in the 1C and 2C cases for three relevant results, together with the corresponding model sizes. 
The Optimized (L) architecture outperforms the baseline (+3.10 and +1.05 percentage points in 1C and 2C, respectively), at the cost of a 1.88$\times$ larger architecture. The Optimized (M) architecture substantially reduces the number of parameters (1,323 vs 3,408), at the price of a slightly lower accuracy than the baseline in the 1C setting, while remaining comparable in the 2C case. The Optimized (S) showcases significantly lower accuracy, yet competitive considering it consists in only 755 parameters.

\begin{table}[]
\centering
\caption{Backbone architecture exploration results after QAT}
\vspace{-0.1cm}
\label{tab:results_bayopt}
\resizebox{\columnwidth}{!}{%
\begin{tabular}{@{}lccc@{}}
\toprule
\textbf{Architecture} & \textbf{N. of parameters} & \textbf{1C Acc. (\%)} & \textbf{2C Acc. (\%)} \\ \midrule
Baseline          & 3,408   (1.00$\times$)&  88.16 ± 6.52        & 67.88 ± 4.29      \\
Optimized (L) & 6,392 (1.88$\times$)& 91.26 ± 3.26        &    68.93 ± 1.70        \\
Optimized (M) & 1,323 (0.39$\times$)&  85.28 ± 4.40       & 68.94 ± 4.19\\
Optimized (S) & 755 (0.22$\times$)& 69.89 ± 4.32        & 61.11 ± 2.32\\
\bottomrule
\end{tabular}%
\vspace{-0.95cm}
}
\end{table}
\vspace{-0.15cm}

\subsection{Latency, memory, and training overhead}

Having demonstrated the effectiveness of CNN+NCM models for IR data, we now analyze the cost of deploying them on MAUPITI. 
We deploy a full-INT8 version of the Optimized (M) model from Tab.~\ref{tab:results_bayopt}, i.e., the one achieving the best accuracy versus size trade-off.
When quantizing the models, the test accuracy drop with respect to floating point versions is equal to 0.47 percentage points for 1C and 1.70 for 2C.

\begin{figure}[t]
    \centering
    \begin{minipage}{0.49\linewidth}
        \centering
        \includegraphics[width=\linewidth]{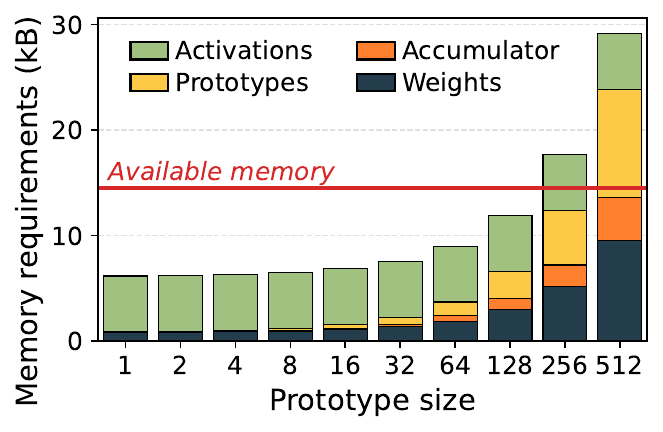}
        \vspace{-0.75cm}
        \caption{Memory requirements of the NCM classifier}
        \label{fig:mem_requirements}
    \end{minipage}
    \hfill
    \begin{minipage}{0.49\linewidth}
    \centering
    \includegraphics[width=\linewidth]{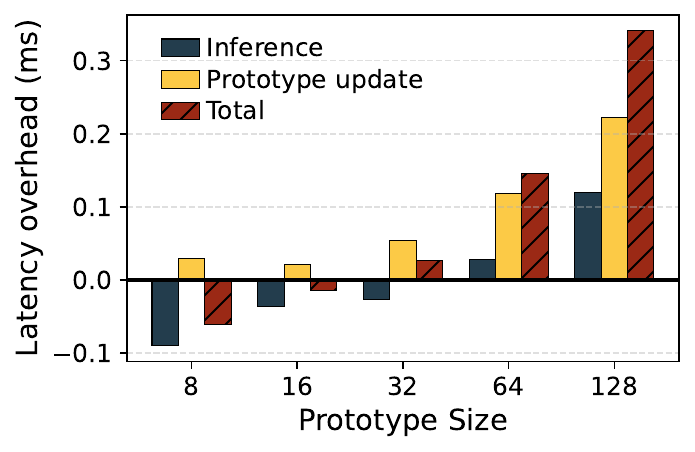}
    \vspace{-0.75cm}
    \caption{Latency overhead with respect to a traditional classifier}
    \label{fig:latenciesps}
    \end{minipage}
    \hfill
    \vspace{-0.3cm}
\end{figure}
Fig.~\ref{fig:mem_requirements} shows the memory occupation breakdown of our method as a function of $p$.
Notably, increasing $p$ affects the weights occupation due to the parameters of the last projection layer. As shown, with $p\le 64$, our model easily fits within the 16\,kB data memory of MAUPITI, with prototypes contributing only a few hundred bytes (maximum $64\times5\times4=1.28$kB where $5$ is the number of classes and $4$ the bytes per element). For comparison, in a 4-shot setting, an input replay buffer would require  $4\times16\times16\times5=5$kB ($4\times$ more), i.e., almost half of the total memory.

We also measure the \textit{inference latency} by executing multiple forward passes and NCM classifications on the real sensor. For $p\le64$ the total classification latency (sensor readout + CNN inference + NCM head) is $\approx 50$\,ms, comfortably below the 100\,ms requirement imposed by the sensor frame rate. The overhead with respect to a traditional classifier (same backbone, 5-neurons softmax head) is $<0.03$\,ms (0.06\% of the total for $p=64$) when considering only the inference phase, as shown in Fig.~\ref{fig:latenciesps}. Notably, inference is slightly \textit{faster} for $p<64$. Closest prototype extraction and the softmax head both require $O(p\cdot C)$ operations, so this difference is caused entirely by non-idealities (e.g. loop overheads). Different distance metrics (Euclidean, Manhattan, angular) yield negligible latency differences.
Lastly, the extra latency overhead due to \textit{prototype updates}, i.e., on-device training, is also $<0.12$\,ms, as expected since each update requires only $O(N_c\cdot p)$ operations per class, making online training effectively real-time. Considering both training and inference, the total overhead sums up to 0.15\,ms (0.29\%).

\subsection{Generalization to other datasets}
\begin{figure}[t]
    \centering
    \includegraphics[width=.9\linewidth]{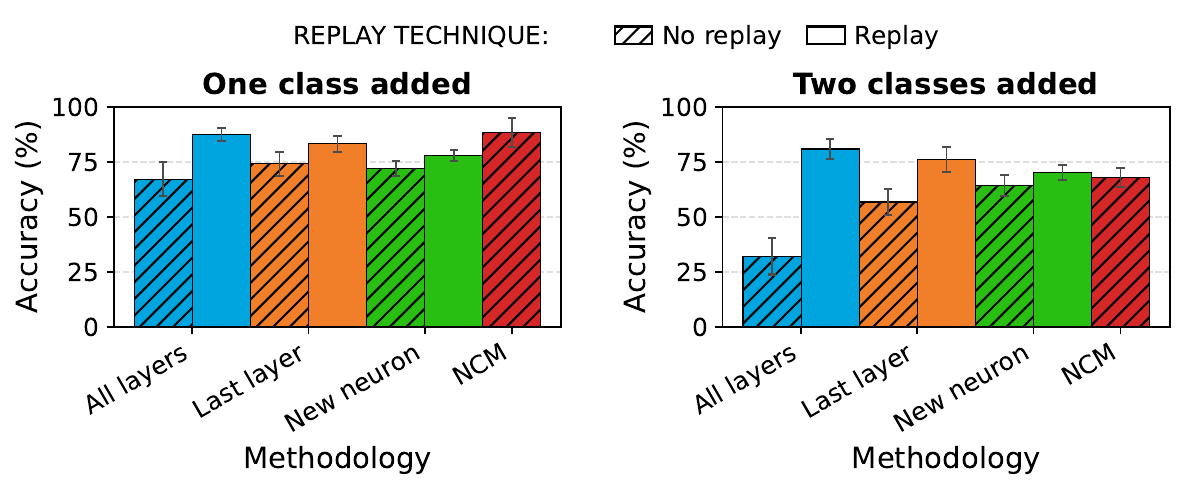}
    \vspace{-0.35cm}
    \caption{9-class accuracy of the NCM classifier and the considered baselines for hand gesture recognition ($p=64$, 4 shots).}
    \label{fig:results_gestures}
    \vspace{-0.38cm}
\end{figure}

To illustrate the generality of our approach, we apply the same CNN+NCM pipeline to the hand gesture recognition task. The encoder architecture and quantization settings are kept identical to those used for pose recognition. We consider $p=64$ and 4-shot training. The results of our method and the baselines after the metric learning training stage are reported in Fig.~\ref{fig:results_gestures}. Despite the lower overall accuracy, given by the higher number of classes and difficulty of the task, the overall trends are confirmed. The NCM achieves similar or better performance than the other non-replay methods, and for the 1C case, remains competitive even with replay, including full-backbone fine-tuning. Notably, these results show that the~sensor is able to autonomously learn 1(2) new gestures, while maintaining $>88.16\%$$(67.88\%)$ accuracy on all 9 classes.

\section{Conclusion}
\label{sec:conclusion}
 In this letter we presented an on-device learning framework based on the MAUPITI smart infrared sensor. We showed how a CNN encoder coupled with a Nearest Class Mean classifier enables efficient and real-time pose and gesture recognition. Particularly, continual learning can be easily applied on-device, without the expensive costs of backpropagation over a quantized model, while matching the performance of a traditional classifier. 

\bibliographystyle{IEEEtran}
\bibliography{bibtex/bib/IEEEexample}
\end{document}